%% file: 1-paper.tex
\documentclass[runningheads]{llncs}
\usepackage{graphicx}
\usepackage{amsmath,amssymb} 
\usepackage{color}
\newcommand*\rot{\rotatebox{90}}
\usepackage{array}
\usepackage{booktabs}
\usepackage{pifont}
\usepackage[width=122mm,left=12mm,paperwidth=146mm,height=193mm,top=12mm,paperheight=217mm]{geometry}

\usepackage[table,xcdraw]{xcolor}
\usepackage{booktabs}
\usepackage{amssymb}
\usepackage{subfig}
\usepackage{xspace}
\newcommand*{\eg}{\emph{e.g.}\@\xspace}
\newcommand*{\ie}{\emph{i.e.}\@\xspace}
\newcommand*{\etal}{\emph{et al.}\@\xspace}
\newcommand*{\etc}{\emph{etc.}\@\xspace}

\begin{document}
\pagestyle{headings}
\mainmatter

\title{Improving Semantic Embedding Consistency by Metric Learning for Zero-Shot Classification} 

\titlerunning{Metric Learning for Zero-Shot Classification}

\authorrunning{Bucher, Herbin and Jurie}

\author{Maxime Bucher$^{1,2}$, St\'ephane Herbin$^1$, Fr\'ed\'eric Jurie$^2$}


\institute{$^1$ONERA - The French Aerospace Lab, Palaiseau, France \\
$^2$Normandie Univ, UNICAEN, ENSICAEN, CNRS}

\maketitle
\input{2-abstract.tex}

\input{3-introduction.tex}

\input{4-relatedwork.tex}

\input{5-method.tex}
\input{6-experiments.tex}

\input{7-conclusions.tex}
\clearpage
\bibliographystyle{splncs}
\bibliography{0-bibfile.bib}

\end{document}

%% file: 2-abstract.tex
\begin{abstract}
This paper addresses the task of zero-shot image classification. The key contribution of the proposed approach is to control the semantic embedding of images -- one of the main ingredients of zero-shot learning -- by formulating it as a  metric learning problem. The optimized empirical criterion associates two types of sub-task constraints: metric discriminating capacity and accurate attribute prediction. This results in a novel expression of zero-shot learning not requiring the notion of class in the training phase: only pairs of image/attributes, augmented with a consistency indicator, are given as ground truth. At test time, the learned model can predict the consistency of a test image with a given set of attributes, allowing flexible ways to produce recognition inferences. Despite its simplicity, the proposed approach gives  state-of-the-art results on four challenging datasets used for zero-shot recognition evaluation.

\keywords{zero-shot learning, attributes, semantic embedding}
\end{abstract}

%% file: 3-introduction.tex
\section{Introduction}

This paper addresses the question of zero-shot learning (ZSL) image classification, \ie, the classification of images belonging to classes not represented by the training examples. This problem has attracted much interest in the last decade because of its clear practical impact: in many applications, having access to annotated data for the categories considered is often difficult, and requires new ways to increase the interpretation capacity of automated recognition systems.The efficiency of ZSL relies on the existence of an intermediate representation level, effortlessly understandable by human designers and sufficiently formal to be the support of algorithmic inferences. Most of the studies  have so far considered this representation in the form of {\em semantic attributes} mainly because it provides an easy way to describe compact yet discriminative descriptions of new classes. 

It has also been observed \cite{Mahajan:2011fg,romera2015embarrassingly} that attribute representations as provided by humans may not be the ideal embedding space because it can lack the informational quality necessary to conduct reliable inferences: the structure of the attribute manifold for a given data distribution may be rather complex, redundant, noisy and unevenly organized. Attribute descriptions, although semantically meaningful and useful to introduce a new category, are not necessarily isomorphic to image data or to image processing outputs.  

To compensate for the shortcomings induced by attribute representations, the recent trends in ZSL studies are aiming at better controlling the classification inference as well as the attribute prediction in the learning criteria. Indeed, if attribute classifiers are learned independently of the final classification task, as in the  Direct Attribute Prediction model~\cite{Lampert:2009ew}, they might be optimal at predicting attributes but not necessarily at predicting novel classes. 

In the work proposed in this paper, we instead suggest that better controlling the structure of the embedding attribute space is at least as important than constraining the classification inference step. The fundamental idea is to empirically disentangle the attribute distribution by learning a metric able to both select and transform the original data distribution according to informational criteria. This metric is obtained by optimizing an objective function based on pairs of attributes/images without assuming that the training images are assigned to categories; only the semantic annotations are used during training. More specifically, we empirically validated the idea that optimizing jointly the attribute embedding and the classification metric, in a multi-objective framework, is what makes the performance better, even with a simple linear embedding and distance to mean attribute classification.

The approach is experimentally validated on 4 recent datasets for zero-shot recognition, \ie the `aPascal\&aYahoo', `Animals with Attributes', `CUB-200-2011 and `SUN attribute' datasets for which excellent results are obtained, despite the simplicity of the  approach.

The rest of the paper is organized as follows: Section~\ref{sec:related} presents the related works, Section~\ref{sec:method} describes the proposed approach while the experimental validation is given in  Section~\ref{sec:expe}.

%% file: 4-relatedwork.tex
\section{Related work\label{sec:related}} 

\subsection{Visual features and semantic attributes}
Image representation -- \ie the set of mechanisms allowing to transform image raw pixel intensities into  representations suitable for recognition tasks -- plays an important role in image classification. State-of-the art image representations were, a couple of years ago, mainly based on the pooling of hard/soft quantized local descriptors (\eg   SIFT \cite{lowe2004distinctive}) through the bag-of-words \cite{csurka2004visual} of Fisher vectors \cite{sanchez2013image} models. However, the work of Krizhevsky et. al. \cite{Krizhevsky:2012wl} has opened a new area and most of the state-of-the-art image descriptors nowadays rely on Deep Convolutional Neural Networks (CNN). We follow this trend in our experiments and use the so-called `VGG-VeryDeep-19' (4096-dim) descriptors of \cite{simonyan2014very}.

Two recent papers have exhibited  existing links between CNN features and semantic attributes. Ozeki  \etal \cite{Ozeki:2014ii} showed that some CNN units can predict some semantic attributes of the `Animals with Attributes' dataset fairly accurately. 
One interesting conclusion of their paper is that the visual semantic attributes can be predicted much more accurately  than the non-visual ones by the nodes of the CNN. 
More recently, \cite{escorcia2015relationship} showed the existence of Attribute Centric Nodes (ACNs) within CNNs trained to recognize objects, collectively encoding information pertinent to visual attributes, unevenly and sparsely distributed across all the layers of the network. 

Despite these recent findings could certainly make the performance of our method better, we don't use them in our experiments and stick to the use of standard CNN features, with the intention of making our results directly comparable to the recent zero-shot learning papers (e.g. \cite{Zhang:2015vs}).  

\subsection{Describing images by semantic and non-semantic attributes}
Zero-shot learning methods rely on the use of intermediate representations, usually given as {\em attributes}. This term can, however, encompass different concepts. For Lampert \etal \cite{Lampert:2014fs}  it denotes the presence/absence of a given object property, assuming that attributes are {\em nameable} properties (color or presence or absence of a certain part, \etc). The advantage of so-defined attributes is that they can be used easily to define new classes expressed by a shared semantic vocabulary. 

However, finding a discriminative and meaningful set of attributes can sometimes be difficult.  \cite{Parikh:2011fg,duan2012discovering} addressed this issue by proposing an interactive approach that discovers local attributes both discriminative and semantically meaningful, employing a recommender system that selects attributes through human interactions. An alternative for identifying attribute vocabulary without human labeling is to mine existing textual description of images sampled from the Internet, such as proposed by \cite{berg2010automatic}. In the same line of thought, \cite{LeiBa:2015vr} presented a model for classifying unseen categories from their (already existing) textual description. 
\cite{Elhoseiny:2013vy} proposed an approach for zero-shot learning  where the description of unseen categories comes in the form of typical text such as an encyclopedia entries, without the need to explicitly define attributes. 

Another drawback of human generated attributes is that they can be redundant or not adapted to image classification. These issues have been addressed by automatically designing discriminative category-level attributes and using them for tasks of cross-category knowledge transfer, such as in the work of Yu \etal \cite{yu2013designing}. Finally, the attributes can also be structured into hierarchies \cite{Verma:2012cu,Rohrbach:2011ue,Akata:2015tv} or obtained by text mining or textual descriptions  \cite{Akata:2015tv,LeiBa:2015vr,Elhoseiny:2013vy}. 

Beside these papers which all consider attributes as meaningful for humans, some authors denoted by ‘attributes’ any latent space providing an intermediate representation between image and hidden descriptions that can be used to transfer information to images from unseen classes. This is typically the case of \cite{wang:2010iu} which jointly learn the attribute classifiers and the attribute vectors, with the intention of obtaining a better attribute-level representation, converting undetectable and redundant attributes into discriminative ones while retaining the useful semantic attributes. \cite{farhadi2009describing} has also introduced the concept of  {\em discriminative attributes}, taking the form of random comparisons. The space of class names can also constitute an interesting embedding, such as in the works of \cite{mensink2014costa,Norouzi:2013tm,Zhang:2015vs,Fu:2015vq} which represent images as mixtures of known classes distributions.  

Finally, it worth pointing out that the aforementioned techniques are restricting attributes to categorical labels and don’t allow the representation of more general semantic relationships. To counter this limitation, \cite{Parikh:2011ge} proposed to model relative attributes by learning a ranking function.

\subsection{Zero shot learning from semantic and attribute embedding}

As defined by \cite{palatucci2009zero}, the problem of zero-shot learning can be seen as the problem of learning a classifier $f : x \rightarrow y$ that can predict novel values of $y$ not available in the training set. Most of existing methods rely on the computation of a similarity or {\em consistency function} linking image descriptors and the semantic description of the classes. These links are given by  learning two embeddings -- the first from the image representation to semantic space and the second from the class space to the semantic space --  and defining a way to describe the constraints between the class space and the image space, the two being strongly interdependent.

DAP (Direct Attribute Prediction) and Indirect attribute prediction (IAP), first proposed by \cite{Lampert:2009ew}, use the between layer of attributes as variables decoupling the images from the layer of labels. In DAP, independent attribute predictors are used to build the embedding of the image, the similarity between two semantic representations (one predicted from the image representation, one given by the class) being given as the probability of the class attribute knowing the image. In IAP, attributes form a connecting layer between two layers of labels, one for classes that are known at training time and one for classes that are not known. In this case, attributes are predicted from (known) class predictions. Lampert \etal \cite{Lampert:2009ew} concluded that DAP gives much better performance than IAP. 

However, as mentioned in the introduction, DAP has two problems: first, it does not model any correlation between attributes, each being predicted independently. Second, the mapping between classes and the attribute space does not weight the relative importance of the attributes nor the correlations between them. 
Inspired by \cite{weston2011wsabie}, which learns a linear embedding between image features and annotations, \cite{Akata:2015cj} and \cite{romera2015embarrassingly} tried to overcome this limitation.
The work of Akata \etal \cite{Akata:2015cj} introduced a function measuring the consistency between an image and a label embedding, the parameters of this function is learned to ensure that, given an image, the correct classes rank higher than the incorrect ones. This consistency function has the form of a bilinear relation $W$ associating the image embedding $\theta(x)$ and the label representation $\phi(y)$ as $S(x,y;W)=\theta(x)^tW\phi(y)$. Romera \etal \cite{romera2015embarrassingly} proposed a simple closed form solution for $W$, assuming a specific form of regularization is chosen. In comparison to our work  none of the two papers (\cite{Akata:2015cj}, \cite{romera2015embarrassingly}) use a metric learning framework to control the statistical structure of the attribute embedding space. 

The coefficients of the consistency constraint $W$ can be also predicted from a semantic textual description of the image. As an example, the goal of \cite{Elhoseiny:2013vy} is to predict a classifier for a new category based only on the learned classes and a textual description of this category.  They solve this problem as a regression function, learnt from the textual feature domain to the visual classifier domain. \cite{LeiBa:2015vr} builds on these ideas, extending them by using a more expressive regression function based on a deep neural network. They take advantage of the architecture of CNNs and learn features at different layers, rather than just learning an embedding space for both modalities. The proposed model provides means to automatically generate a list of pseudo-attributes for each visual category consisting of words from Wikipedia articles.

In contrast with the aforementioned methods, Hamm et al. \cite{Hamm:2015uw} introduced the idea of ordinal similarity between classes (eg. d(`cat',`dog') $<$ d(`cat', `automobile')), claiming that not only this type of similarity may be sufficient for distinguishing cat and truck, but also that it seems a more natural representation since the ordinal similarity is invariant under scaling and monotonic transformation of numerical values. It is also worth mentioning the work of Jayaraman \etal  \cite{Jayaraman:2014wq} which proposed to leverage the statistics about each attribute error tendencies within a random forest approach, allowing to train zero-shot models that explicitly account for the unreliability of attribute predictions.   

Wu \etal \cite{Wu:2014tp}  exploit natural language processing technologies to generate event descriptions. They measure their similarity to images by projecting them into a common high-dimensional space using text expansion. The similarity is expressed as the concatenation of $L_2$ distances of the different modalities considered.  Strictly speaking, there is no metric learning involved but a concatenation of $L_2$ distances. Finally, Frome \etal \cite{Frome:2013ux} aim at leveraging semantic knowledge learned in the text domain, and transfer it to a model trained for visual object recognition by learning a metric aligning the two modalities. However, in contrast to our work, \cite{Frome:2013ux,Wu:2014tp} do not explicitly control the quality of the embedding.


\subsection{Zero-shot learning as transductive and semi-supervised learning}
 
All the previously mentioned approaches consider that the embedding and the consistency function have to be learned from a set of training data of known classes, used in a second time to infer predictions about the images of new classes not available during training. However, a different problem can be addressed when images from the unknown classes are already available at training time and can hence be used to produce a better embedding. In this case, the problem can be cast as a transductive learning problem \ie the inference of the correct labels for the given unlabeled data only, or a semi-supervised learning  problem \ie the inference of the best embedding using both labeled and unlabeled data.

Wang and Forsyth  \cite{Wang:2009wm} proposed MIL framework for jointly learning attributes and object classifiers from weakly annotated data. Wang and Mori \cite{wang:2010iu}  treated attributes of an object as latent variables and captured the correlations among attributes using an undirected graphical model,  allowing to infer object class labels using the information of both the test image and its (latent) attributes. In \cite{Mahajan:2011fg},  the class information is incorporated into the attribute classifier to get an attribute-level representation that generalizes well to unseen examples of known classes as well as those of the unseen classes, assuming unlabeled images are available for learning. \cite{Socher:2013vb} considered the introduction of unseen classes as a novelty detection problem in a multi-class classification problem. If the image is of a known category, a standard classifier can be used. Otherwise, images are assigned to a class based on the likelihood of being an unseen category.  Fu \etal \cite{fu2014transductive} rectified the projection domain shift between auxiliary and target datasets by introducing a multi-view semantic space alignment process to correlate different semantic views and the low-level feature view, by projecting them onto a latent embedding space learnt using multi-view Canonical Correlation Analysis. More recently Li \etal \cite{Li:2015wq} learned the embedding from the input data in a semi-supervised large-margin learning framework, jointly considering multi-class classification over observed and unseen classes. Finally, \cite{Kodirov:2015vv}  formulated a  regularized sparse coding framework which used the target domain class label projections in the semantic space to regularize the learnt target domain projection, with the aim of overcoming the projection domain shift problem.  

\subsection{Zero-shot learning as a metric learning problem}
Two contributions, \cite{mensink2012metric} and \cite{kuznetsova2016}, exploit metric learning to Zero shot class description. Mensink \etal \cite{mensink2012metric} learn a metric adapted to  measure the similarity of images, in the context of $k$-nearest neighbor image classification, and apply it in fact to One Shot Learning to show it can generalize well to new classes. They don't use any attribute embedding space nor consider ZSL in their work. Kuznetsova \etal \cite{kuznetsova2016} learn a metric to infer pose and object class from a single image. They use the expression \emph{zero-shot} to actually denote a (new) transfer learning problem when data are unevenly sampled in the joint pose and class space, and not a Zero-Shot Classification problem where new classes are only known from attribute descriptions.

As far as we know, zero-shot learning has never been addressed explicitly as a metric learning problem in the attribute embedding space, which is one of the key contributions of this paper.


%% file: 5-method.tex
\section{Method\label{sec:method}}

\subsection{Embedding consistency score}

Most of the inference problems can be cast into an optimal framework of the form:
\[ \mathbf{Y}^* = \operatorname*{arg\,min}_{\mathbf{Y}\in\mathcal{Y}} S(\mathbf{X},\mathbf{Y}) \]
where $\mathbf{X}\in \mathcal{X}$ is a given sample from some modality, \eg an image or some features extracted from it, $\mathbf{Y}^*$ is the most consistent association from another modality $\mathcal{Y}$, \eg a vector of attribute indicators or a textual description, and  $S$ is a measure able to quantify the joint consistency of two observations from the two modalities. In this formulation, the smaller the score, the more consistent the samples. One can think of this score as a negative likelihood.

When trying to design such a consistency score, one of the difficult aspects is to relate meaningfully the two modalities. One usual approach consists in embedding them into a common representational space $\mathcal{A}$\footnote{We use the letters $A$ and $\mathcal{A}$ in our notations since we will focus on the space of \emph{attribute} descriptions as the embedding space.} where their heterogeneous nature can be compared.  This space can be abstract, \ie its structure can be obtained from some optimization process, or semantically interpretable \eg a fixed list of attributes or properties each indexed by a tag referring to some shared knowledge or ontology, leading to a $p$-dimensional vector space. Let $\hat{\mathbf{A}}_X(\mathbf{X})$ and $\hat{\mathbf{A}}_Y(\mathbf{Y})$ be the two embeddings for each modality $X$ and $Y$, taking values in $\mathcal{X}$ and $\mathcal{Y}$ and producing outputs in $\mathcal{A}$.
 
In this work, it is proposed to define the consistency score as a metric on the  common embedding space $\mathcal{A}$. More precisely, we use the Mahalanobis like description of a metric parametrized by a linear mapping $\mathbf{W}_A$: 
\[
d_A(\mathbf{A}_1,\mathbf{A}_2) = \left\Vert (\mathbf{A}_1 - \mathbf{A}_2)^T\mathbf{W}_A \right\Vert_2,
\label{eq:embedding}
\]
assuming that the embedding space is a vector space, and define the consistency score as:

\begin{equation}
S(\mathbf{X},\mathbf{Y}) = d_A(\hat{\mathbf{A}}_X(\mathbf{X}), \hat{\mathbf{A}}_Y(\mathbf{Y})) \nonumber\\
= \left\Vert(\hat{\mathbf{A}}_X(\mathbf{X}) - \hat{\mathbf{A}}_Y(\mathbf{Y}))^T \mathbf{W}_A \right\Vert_2.
\label{eq:score}
\end{equation}

The Mahalanobis mapping $\mathbf{W}_A$ can be interpreted itself as a linear embedding in an abstract $m$-dimensional vector space where the natural metric is the Euclidean distance, and acts as a multivariate whitening filter. It is expected that this property will improve empirically the reliability of the consistency score~(\ref{eq:score}) by choosing the appropriate linear mapping.

We are now left with two questions: how to define the embedding? How build the Mahalanobis mapping? We see in the following that these two questions can be solved jointly by optimizing a unique criterion.

\subsection{Embedding in the attribute space}

The main problem addressed in this work is to be able to discriminate a series of new hypotheses that can only be specified using a single modality, the $Y$ one with our notations. In many Zero-Shot Learning studies, this modality is often expressed as the existence or presence of several attributes or properties from a fixed given set. The simplest embedding space one can think of is precisely this attribute space, implying that the $Y$ modality embedding is the identity: $\hat{\mathbf{A}}_Y(\mathbf{Y}) = \mathbf{Y}$ with $\mathcal{A}= \mathcal{Y}$. In this case, the consistency score simplifies as:
\begin{equation}
S(\mathbf{X},\mathbf{Y}) = \left\Vert(\hat{\mathbf{A}}_X(\mathbf{X}) - \mathbf{Y})^T \mathbf{W}_A \right\Vert_2
\label{eq:score_attribute_embedding}
\end{equation}

The next step is to embed the $X$ modality into $Y$ directly. We suggest using a simple linear embedding with matrix $\mathbf{W}_X$ and bias $\mathbf{b}_X$, assuming that $X$ is in a $d$-dimensional vector space. This can be expressed as:
\begin{equation}
\hat{\mathbf{A}}_X(\mathbf{X}) = \max(0,\mathbf{X}^T\mathbf{W}_X  + \mathbf{b}_X).
\label{eq:embedding_attribute}
\end{equation}
We use a reLu-type output normalization to keep the significance of the attribute space as property detectors, negative numbers being difficult to interpret in this context.

In the simple formulation proposed here, we do not question the way new hypotheses are specified in the target modality, nor use any external source of information (\eg word vectors) to map the attributes into a more semantically organized space such as in \cite{Socher:2013vb}. We leave the problem of correcting the original attribute description to the construction of the metric in the common embedding space.

\subsection{Metric learning}

The design problem is now reduced to the estimation of three mathematical objects: the linear embedding to the attribute space $\mathbf{W}_X$ of dimensions $d\times p$, a bias $\mathbf{b}_X$ of dimension $p$, and the Mahalanobis linear mapping $\mathbf{W}_A$ of dimensions $p\times m$,  $m$ being a free parameter to choose.

The proposed approach consists in building empirically those objects from a set of examples by appling metric learning techniques.   The training set is supposed to  contain pairs of data $(\mathbf{X}_i,\mathbf{Y}_i)$ sampling the joint distribution of the two modalities: $\mathbf{X}_i$ is a vector representing an image or some features extracted from it, while $\mathbf{Y}_i$ denotes an attribute-based description. Notice that we do not introduce any class information in this formulation: the link between class and attribute representations is assumed to be specified by the use case considered.

The rationale behind the use of metric learning is to transform the original representational space so that the resulting metric takes into account the statistical structure of the data using pairwise constraints. One usual way to do so is to express the problem as a binary classification on pairs of samples, where the role of the metric is to separate similar and dissimilar samples by thresholding (see ~\cite{Bellet:2013uv} for a survey on M.L.). It is easy to build pairs of similar and dissimilar examples from the annotated examples by sampling randomly (uniformly or according to some law) the two modalities $\mathcal{X}$ and $\mathcal{Y}$ and assigning an indicator $Z\in \{-1,1\}$ stating whether $\mathbf{Y}_i$ is a good attribute description of $\mathbf{X}_i$ ($Z_i=1$) or not ($Z_i=-1$). Metric learning approaches try to catch a data-dependent way to encode similarity. In general, the data manifold has a smaller intrinsic dimension than the feature space, and is not isotropically distributed.

We are now given a dataset of triplets $\{(\mathbf{X}_i,\mathbf{Y}_i, Z_i)\}_{i=1}^N$, the $Z$ indicator stating that the two modalities are similar, \ie consistent, or not\footnote{To make notations simpler, we do not rename or re-index from the original dataset the pairs of data for the similar and dissimilar cases.}. The next step is to describe an empirical criterion that will be able to learn $\mathbf{W}_X$, $\mathbf{b}_X$ and $\mathbf{W}_A$. The idea is to decompose the problem in three objectives: metric learning, good embedding and regularization.

The metric learning part follows a now standard hinge loss approach \cite{shalev2004online} taking the following form for each sample:
\begin{equation}
l_H(\mathbf{X}_i,\mathbf{Y}_i,Z_i,\tau) = \max\left(0,1-Z_i(\tau - S(\mathbf{X}_i,\mathbf{Y}_i)^2)\right).
\label{eq:criterion_hinge_loss}
\end{equation}
The extra parameter $\tau$ is free and can also be learned from data. Its role is to define the threshold separating similar from dissimilar examples, and should depend on the data distribution.

The embedding criterion is a simple quadratic loss, but only applied to similar data:
\begin{equation}
l_A(\mathbf{X}_i,\mathbf{Y}_i,Z) = \max(0,Z_i).\left\Vert \mathbf{Y}_i - \hat{\mathbf{A}}_X(\mathbf{X}_i) \right\Vert_2^2.
\label{eq:criterion_attribute_prediction}
\end{equation}
Its role is to ensure that the attribute prediction is of good quality, so that the difference $\mathbf{Y} - \hat{\mathbf{A}}_X(\mathbf{X})$ reflects dissimilarity due to modality inconsistencies rather than bad representational issues.

The size of the learning problem ($d\times p + p + p\times m$) can be large and requires regularization to prevent over fitting. We use a quadratic penalization: 
\begin{equation}
R(\mathbf{W}_A,\mathbf{W}_X, \mathbf{b}_X) = \left\Vert \mathbf{W}_X \right\Vert_F^2 + \left\Vert \mathbf{b}_X \right\Vert_2^2 + \left\Vert \mathbf{W}_A \right\Vert_F^2 
\label{eq:criterion_weight_regularization}
\end{equation}
where $\left\Vert . \right\Vert_F$ is the Frobenius norm.

The overall optimization criterion can now be written as the sum of the previously defined terms:
\begin{equation}
\begin{split}
\mathcal{L}(\mathbf{W}_A,\mathbf{W}_X, \mathbf{b}_X, \tau) = \sum_i l_H(\mathbf{X_i},\mathbf{Y_i},Z_i,\tau) + \lambda \sum_i l_A(\mathbf{X_i},\mathbf{Y_i},Z_i)\\
 + \mu R(\mathbf{W}_A,\mathbf{W}_X, \mathbf{b}_X)
\label{eq:criterion_global}
\end{split}
\end{equation}
where $\lambda$ and $\mu$ are hyper-parameters that are chosen using cross-validation. Note that the criterion~(\ref{eq:criterion_global}) can also be interpreted as a multi-objective learning approach since it mixes two optimal but dependent issues: attribute embedding and metric on the embedding space.

To solve the optimization problem, we do not follow the approach proposed in \cite{shalev2004online} since we also learn the attribute embedding part $\mathbf{W}_X$ jointly with the metric embedding $\mathbf{W}_A$. We use instead a global stochastic gradient descent (see section~\ref{sec:expe} for details).

\subsection{Application to image recognition and retrieval}

The consistency score~(\ref{eq:score_attribute_embedding}) is a versatile tool that can be used for several image interpretation problems. Section~\ref{sec:expe} will evaluate the potential of our approach on three of them.

\vspace{-.5em}\subsubsection*{Zero-shot learning}
The problem can be defined as finding the most consistent attribute description given the image to classify, and  a set of exclusive attribute class descriptors $\{\mathbf{Y}^*_k\}_{k=1}^C$ where $k$ is the index of a class:

\begin{equation}
k^* = \operatorname*{arg\,min}_{k\in\{1\ldots C\}} S(\mathbf{X},\mathbf{Y}^*_k)
\label{eq:zero_shot_classification}
\end{equation}

In this formulation, classifying is made equivalent to identifying between the $C$ classes the best attribute description. A variant of this scheme can exploit a voting process to identify the best attribute among a set of $k$ candidates, inspired from a $k-$nearest neighbor approach. 

\vspace{-.5em}\subsubsection*{Few-shot learning}

Learning a metric in the embedding space can conveniently be used to specialize the consistency score to new data when they are available. We study a simple fine tuning approach using stochastic gradient descent on criterion~(\ref{eq:criterion_global}) applied to novel triplets $(X,Y,Z)$ from unseen classes only, starting with the model learned with seen classes. This makes \textquotedblleft few-shot learning\textquotedblright possible. The decision framework is identical to the ZSL one.

\vspace{-.5em}\subsubsection*{Zero-shot retrieval}
The score~(\ref{eq:score_attribute_embedding}) can also used to retrieve the data from a given database that have at least a consistent level $\lambda$  with a given query defined in the $Y$ (or $A$) modality:

\[
\mathrm{Retrieve}(\mathbf{A},\lambda) = \{\mathbf{X}\in\mathcal{X} \,/\, S(\mathbf{X},\mathbf{A}) < \lambda\}
\]

The performance is usually characterized  by precision-recall curves.

%% file: 6-experiments.tex
\section{Experiments\label{sec:expe}}
This section presents the experimental validation of the proposed method. The section first introduces the 4 datasets evaluated as well as the details of the experimental settings. The method is empirically evaluated  on three different tasks as described in section~\ref{sec:method}: Zero-Shot-Learning (ZSL), Few-Shot Learning (FSL) and Zero-Shot Retrieval (ZSR). The ZSL experiments aim at evaluating the capability of the proposed model to predict unseen classes. This section also evaluates the contribution of the different components of the model to the performance, and makes comparisons with state-of-the-art results. In the FSL experiments, we show how the ZSL model can serve as good prior to learning a classifier when only a few samples of the unknown classes are available. Finally, we evaluate our model on a ZSR task, illustrating the capability of the algorithm to retrieve images using  attribute-based queries.  

\subsection{Datasets and Experimental Settings}

The experimental valuation is done on 4 public datasets widely used in the community, allowing to compare our results with those recently proposed in the literature: the aPascal\&aYahoo (aP\&Y) \cite{farhadi2009describing}, Animals with Attributes (AwA) \cite{Lampert:2009ew}, CUB-200-2011 (CUB) \cite{Wah:2011vq} and SUN attribute (SUN) \cite{Patterson:2014cv} datasets (see Table~\ref{statsdata} for few statistics on their content).
Theses datasets exhibit a large number of categories (indoor and outdoor scenes, objects, person, animals, \etc) and attributes (shapes, materials, color, parts, \etc)

\begin{table}[tb]
\centering
\caption{Dataset statistics}
\label{statsdata}
\begin{tabular}{|l|c|c|c|c|}
\hline
Dataset &\#Training classes & \#Test classes&\#Instances& \#Attributes \\\hline\hline
aPascal \& aYahoo \cite{farhadi2009describing}& 20 & 12 & 15,339 & 64 \\\hline
Animals with Attributes \cite{Lampert:2009ew}& 40 & 10 & 30,475 & 85 \\\hline
CUB 200-2011 \cite{Wah:2011vq}& 150 & 50 & 11,788 & 312 \\\hline
SUN Attributes \cite{Patterson:2014cv}& 707 & 10 & 14,340 & 102 \\\hline
\end{tabular}
\end{table}

These datasets have been introduced for training and evaluating ZSL methods and contain images annotated with semantic attributes. More specifically, each image of the aP\&Y, CUB and SUN datasets has its own attribute description, meaning that two images of the same class can have different attributes. This is not the case for AwA where all the images of a given class share the same attributes. As a consequence, in the ZSL experiments on aP\&Y, CUB and SUN, the attribute representation of unknown classes, required for class prediction, is taken as their mean attribute frequencies. 

In order to make comparisons with previous works possible, we use the same training/testing splits as  \cite{farhadi2009describing} (aP\&Y), \cite{Lampert:2009ew} (AwA), \cite{Akata:2015tv} CUB and  \cite{Jayaraman:2014wq} (SUN). 

Regarding the representation of images, we used both the VGG-VeryDeep-19  \cite{simonyan2014very} and   AlexNet \cite{Krizhevsky:2012wl} CNN models, both pre-trained on imageNet -- without fine tuning to the attribute datasets -- and use the penultimate fully connected layer (\eg, FC7 4096-d layer for VGG-VeryDeep-19) for representing the images.  Very deep CNN models act as generic feature extractors and have been demonstrated to work well for object recognition. They have been also used in many recent ZSL experiments and we use exactly the same descriptors as  \cite{Akata:2015tv,Zhang:2015vs}. 

One of the key characteristics of our model is that it requires a set of image/attributes  pairs for training. Positive (resp. negative) pairs are obtained by taking  the training images associated with their own provided attribute vector (resp. by  randomly assigning  attributes not present in the image) and are assigned to the class label `1’ (resp. `-1’). In order to bound the size of the training set we generate only 2 pairs per training image, one positive and one negative.

Our model has three hyper-parameters:  the weight $\lambda$, the dimensionality of the space in which the distance is computed ($m$) and the regularization parameters $\mu$. These  hyper-parameters are estimated through a grid search validation procedure by randomly keeping 20\% of the training classes for cross-validating the hyper-parameters, and choosing the parameters giving best accuracy for these so-obtained validation classes.  The parameter are searched in the following ranges: $m \in$ $[20\%, 120\%]$ of the initial attribute dimension, $\lambda \in$  $[0.05,1.0]$ and $\mu \in$ $[0.01,10.0$.  $\tau$ is a parameter learned during training. 

 Once the hyper-parameters are tuned, we take the whole training set to learn the final model and evaluate it on the test set (unseen classes in case of ZSL).  

The optimization of $\mathbf{W}_A$ and $\mathbf{W}_X$ is done with  stochastic gradient descent, the parameters being initialized randomly with normal distribution. The size of the mini-batch is of 100. As the objective function is non-convex, different initializations can give different parameters. We addressed this issue by doing 5 estimations of the parameters starting from 5 different initializations and selecting the best one on a validation set (we keep a part of the train set for doing this and fine-tune the parameters on the whole train set when the best initialization is known).
We use the optimizer provided in the TensorFlow framework \cite{tensorflow2015-whitepaper}.  Using the GPU mode with a Nvidia 750 GTX GPU, learning a model ($\mathbf{W}_A$ and $\mathbf{W}_X$) takes 5-10 minutes for a given set of hyper-parameters. Computing image/attribute consistency takes around 4ms per pair.

\subsection{Zero-Shot Learning Experiments}
The experiments follow the standard ZSL protocol: during training, a set of images from known classes is available for learning the model parameters. At test time, images from unseen classes are processed and the goal is to find the class described by an attribute representation most consistent with the images. 

Table \ref{acczstab} gives the performance  of our approach on the 4 datasets considered -- expressed as multi-class accuracy -- and makes comparisons with state-of-the-art approaches. The performances of previous methods are taken from \cite{Zhang:2015vs,Akata:2015tv,zhang2016zero}. Performance is reported with 2 different features \ie VGG-VeryDeep-19 \cite{simonyan2014very} and  AlexNet \cite{Krizhevsky:2012wl} for fair comparisons. As images of AwA are not public anymore it is only possible to use the features available for download. On the four datasets our model achieves above state-of-the-art performance (note : \cite{zhang2016zero} was published after our submission), with a noticeable improvement of more than 8\% on aP\&Y.

\begin{table}[tb]
\centering
\caption{Zero-shot classification accuracy (mean $\pm$ std). We report results both with VGG-verydeep-19 \cite{simonyan2014very} and AlexNet \cite{Krizhevsky:2012wl} features for fair comparisons, whenever it's possible (AwA images are not public anymore preventing the computation of their AlexNet representations).}
\label{acczstab}
\begin{tabular}{c|l|c|c|c|c|}
\hline
Feat.& Method & {aP\&Y} & {AwA} & {CUB} & {SUN}\\ \hline\hline 
&Akata \etal \cite{Akata:2015tv} & - & 61.9 & 40.3 & - \\
 \rot{\rlap{\small Alex}} \rot{\rlap{\small Net}} \rot{\rlap{\cite{Krizhevsky:2012wl}}}&Ours   & 46.14$\pm$0.91 & - &  {\bf 41.98$\pm$0.67} & 75.48$\pm0.43$ \\ \hline \hline 
&Lampert \etal \cite{Lampert:2014fs} & 38.16 & 57.23 & - & 72.00 \\
&Romera-Paredes \etal \cite{romera2015embarrassingly}  & 24.22$\pm2.89$ & 75.32$\pm2.28$ & - & 82.10$\pm0.32$ \\
&Zhang \etal \cite{Zhang:2015vs}  & 46.23$\pm0.53$ & 76.33$\pm0.83$ & 30.41$\pm0.20$ & 82.50$\pm$1.32\\
&Zhang \etal \cite{zhang2016zero}  & 50.35$\pm2.97$ & {\bf 80.46$ \bf \pm0.53$} & 42.11$\pm0.55$ & 83.83$\pm$0.29\\ \cline{2-6}
&Ours w/o ML & {47.25$\pm0.48$} & {73.81$\pm0.13$} & {33.87$\pm0.98$} & 74.91$\pm0.12$ \\ 
&Ours w/o constraint & {48.47$\pm1.24$} & {75.69$\pm0.56$} & {38.35$\pm0.49$} & 79.21$\pm0.87$ \\\cline{2-6} 
\rot{\rlap{\small  VGG-VeryDeep}}\rot{\rlap{~~~~~~~\cite{simonyan2014very}}}&Ours &{\bf 53.15$\pm$0.88} & { 77.32$\pm$1.03} & {\bf 43.29$\pm$0.38} & {\bf 84.41$\pm$0.71} \\ \hline
\end{tabular}
\end{table}

As explained in the previous section, our model is based on a multi-objective function trying to maximize metric discriminating capacity as well as attribute prediction. It is interesting to observe how the performance degrades when one of the two terms is missing. In Table \ref{acczstab}, the `Ours w/o ML' setting makes use of the Euclidean distance \ie $\mathbf{W}_A=\mathbb{I}$.  The `Ours w/o constraint' setting is when the attribute prediction term (Eq.~\ref{eq:criterion_attribute_prediction}) is missing in the criterion. This term gives a 4\% improvement, on average.


Figure~\ref{fig:accdim} shows the accuracy as a function of the embedding dimension. This projection maps the original data in a space in which the Euclidean distance is good for the task considered. It can be seen as a way to exploit and select the correlation structure between attributes. We experimented that the best performance is generally obtained when the dimension of this space less than 40\% smaller than the size of the initial attribute space.

\subsection{Few-Shot Learning}


\begin{figure}[tb]
    \centering
    \subfloat[Classification accuracy]{\label{fig:accdim}{\includegraphics[width=0.46\textwidth]{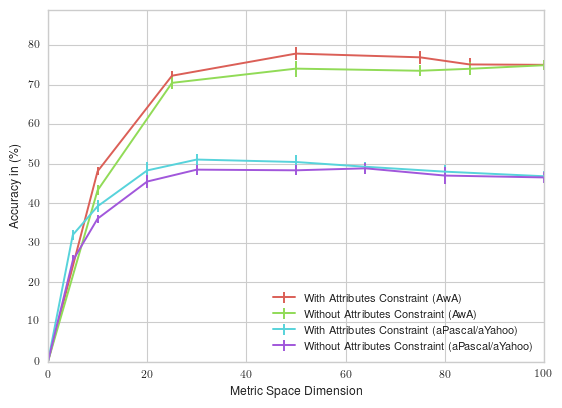} }}%
    \qquad
    \subfloat[Image retrieval]{\label{fig:fewshot}{\includegraphics[width=0.46\textwidth]{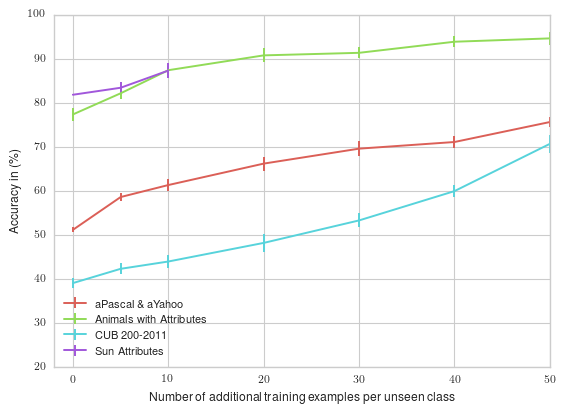} }}%
      \caption{\protect\subref{fig:accdim} ZSL accuracy as a function of the dimensionality of the metric space; best results are obtained when the dimension of the metric embedding is less than 40\% of the image space dimension. It also shows the improvement due to the attribute prediction term in the objective function. \protect\subref{fig:fewshot} Few-shot learning: Classification accuracy (\%) as a function of the amount of  training examples per unseen classes.}%
\end{figure}

Few-shot learning corresponds to the situation where 1 (or more)  annotated example(s) from unseen classes are  available at test time. In this case our model is first trained using only the seen classes (same as with ZSL), and we introduced the examples from unseen class data one by one before fine-tuning the model parameters by doing a few more learning iterations using these new data only. 

Figure~\ref{fig:fewshot} shows the accuracy evolution, given as a  function of the number of additional images from the unseen classes. Please note that for the SUN dataset we have used a maximum of 10 additional examples as unseen classes contain only 20 images. We observed that knowing even a very few number of annotated examples significantly improves the performance. It is a very encouraging behavior for large-scale applications where annotations for a large number of categories are hard and expensive to get.

\subsection{Zero-Shot Retrieval}
The task of Zero-Shot image Retrieval consists in searching an image database with attribute-based queries. For doing this, we first train our model as for standard  ZSL. We then take the attribute descriptions of unseen classes as queries, and rank the images from the unseen classes based on the similarity with the query. Table~\ref{zsret} reports the mean average precision  on the 4 datasets. Our model outperforms the state-of-the-art SEE method~\cite{Zhang:2015vs} by more than 10\% on average. 

\begin{table}[htb]
\vspace{-1em}\centering
\caption{Zero-Shot Retrieval task: Mean Average Precision (\%)  on the 4 datasets}
\label{zsret}
\begin{tabular}{|l|c|c|c|c|c|}
\hline
       & {aP\&Y} & {AwA} & {CUB} & SUN & Av.  \\
 \hline
Zhang \etal \cite{Zhang:2015vs} &15.43&46.25& 4.69& \bf 58.94& 31.33\\ \hline
Ours (VGG features)& \bf 36.92&\bf 68.1 &\bf 25.33        & 52.68 & \bf 45.76 \\ \hline
\end{tabular}
\end{table}

\begin{figure}[tb]
    \centering
    \subfloat[aP\&Y]{{\includegraphics[width=0.46\textwidth]{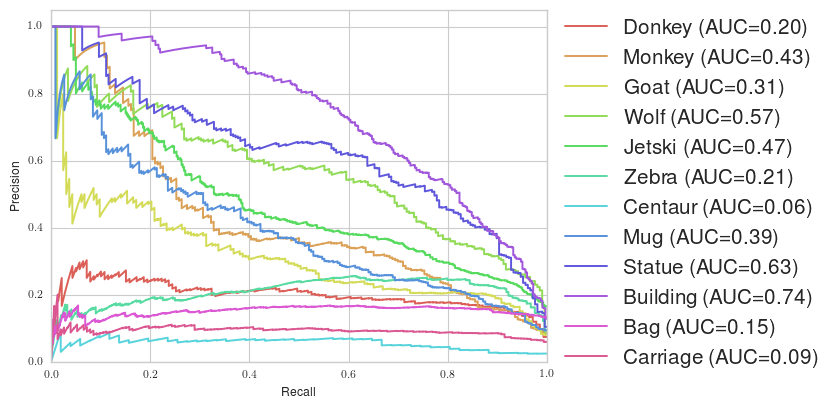} }}%
    \qquad
    \subfloat[AwA]{{\includegraphics[width=0.46\textwidth]{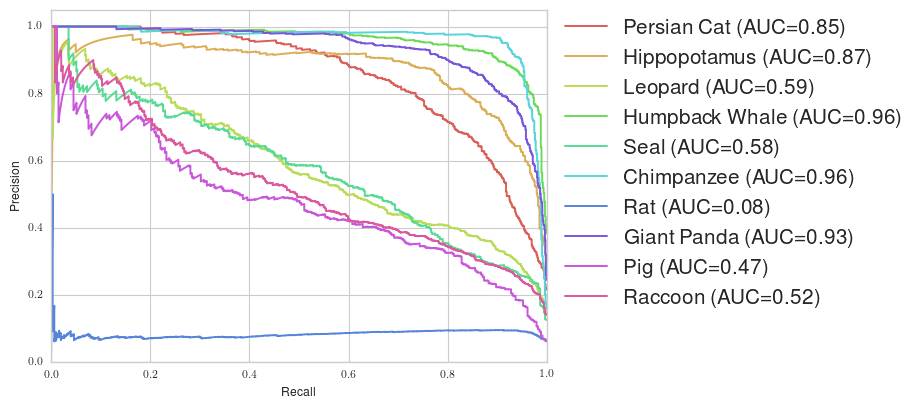} }}%
    \qquad
    \subfloat[CUB-200-2011]{{\includegraphics[width=0.46\textwidth]{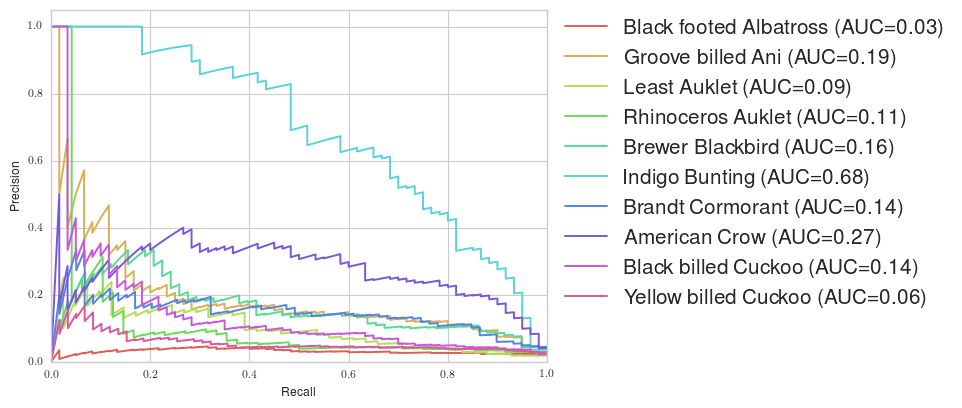} }}%
    \qquad
    \subfloat[SUN]{{\includegraphics[width=0.46\textwidth]{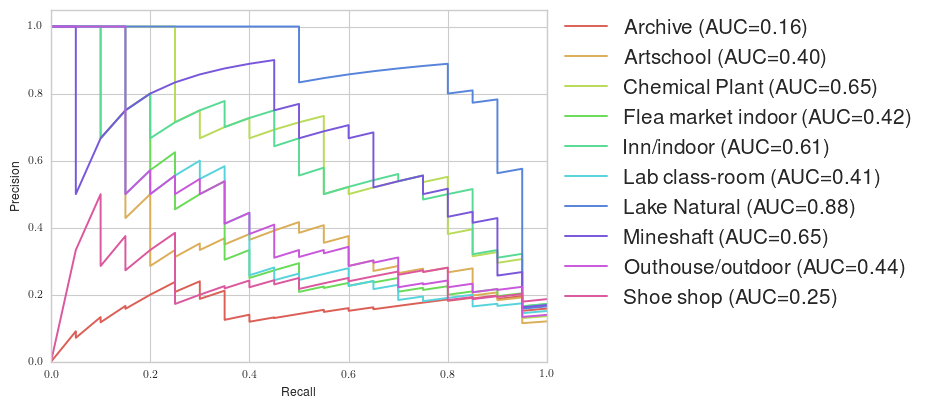} }}%
      \caption{Precision Recall curve for each unseen class by dataset. For CUB dataset we randomly choose  10 classes (best viewed on a computer screen).}%
    \label{fig:presrec}%
\end{figure}

Figure~\ref{fig:presrec} shows the average precision for each class of the 4 datasets. In the aP\&aYa dataset, the `donkey', `centaur' and `zebra' classes have a very low average precision. This can be explained by the strong visual similarity between these classes which only differ by a few attributes.

%% file: 7-conclusions.tex
\vspace{-.3cm}
\section{Conclusion}
\vspace{-.3cm}

This paper has presented a novel approach for zero-shot classification exploiting multi-objective metric learning techniques. The proposed formulation has the nice property of not requiring any ground truth at the category level for learning a consistency score between the image and the semantic modalities, but only requiring weak consistency information. The resulting score can be used with versatility on various image interpretation tasks, and shows close or above state-of-the-art performance on four standard benchmarks. The formal simplicity of the approach allows several avenues for future improvement. A first one would be to provide a better embedding on the semantic side of the consistency score $\hat{\mathbf{A}}_Y(\mathbf{Y})$. A second one would be to explore more complex functions than the linear mappings tested in this work, and introduce, for instance, deep network architectures.